\documentclass[letterpaper, 10 pt, conference]{ieeeconf}  

\IEEEoverridecommandlockouts                              
                                                          
\usepackage{times}

\usepackage{multicol}
\usepackage{caption,subcaption,tabularx,booktabs}
\newcolumntype{?}{!{\vrule width 0.2mm}}  
\usepackage{array}
\usepackage[bookmarks=true]{hyperref}
\usepackage{graphicx}
\usepackage{amsmath}
\usepackage{amssymb}
\usepackage{bm}
\usepackage{acronym}
\usepackage{siunitx}

\usepackage{tikz} 
\usetikzlibrary{calc, arrows, shapes, positioning, decorations.pathreplacing, shadings, angles}
\pgfdeclareverticalshading{rainbow}{100bp}
{color(0bp)=(magenta);
 color(25bp)=(magenta);
 color(35bp)=(blue);
 color(45bp)=(cyan);
 color(55bp)=(green);
 color(65bp)=(yellow);
 color(75bp)=(red);
 color(100bp)=(red)}

\usepackage{lipsum}

\usepackage{xspace}

\newcommand{\refappendix}[1]{\hyperref[#1]{Appendix~\ref*{#1}}}

\acrodef{FMCW}{Frequency-Modulated Continuous Wave}
\acrodef{GP}{Gaussian process}
\acrodef{ICP}{Iterative Closest Point}
\acrodef{IMU}{Inertial Measurement Unit}
\acrodef{WNOA}{White-Noise-on-Acceleration}
\acrodef{MAP}{Maximum A Posteriori}
\acrodef{IMU}{Inertial Measurement Unit}
\acrodef{DOF}{degrees-of-freedom}
\acrodef{2D}{two-dimensional}
\acrodef{3D}{three-dimensional}

\newcommand{\mbf}{\mathbf}
\newcommand{\mbs}[1]{{\boldsymbol{#1}}}

\newcommand{\I}{\bm{1}}
\newcommand{\gT}[1]{\mbs{\mathcal{T}}_{\!\!#1}}
\newcommand{\gC}[1]{\bm{C}_{#1}}
\newcommand{\gR}[1]{\bm{R}_{#1}}
\newcommand{\gr}[2]{{\bm{t}_{#1}^{#2}}}
\newcommand{\gvp}[2]{\mbs{\varpi}_{#1}^{#2}}
\newcommand{\gqh}[2]{{{}\hat{\mbf{q}}_{#1}^{#2}}}
\newcommand{\gq}{\mbf{q}}
\newcommand{\gqi}{\mbf{q}^{ij}_j}

\newcommand{\bA}{\bm{A}}
\newcommand{\bB}{\bm{B}}
\newcommand{\bx}{\bm{x}}
\newcommand{\by}{\bm{y}}

\newcommand{\norm}[1]{ \lVert #1 \rVert }
\DeclareMathOperator{\nullspacenp}{null}
\newcommand{\nullspace}[1]{ \nullspacenp\left( #1 \right)}

\newtheorem{lemma}{Lemma}

\newcommand{\Dw}{\mbf{D}}
\newcommand{\bv}{\mbs{\varpi}}
\newcommand{\adT}{\mbs{\mathcal{T}}}
\newcommand{\edop}{e_{\text{dop}}}
\newcommand{\ydop}{y_{\text{dop}}}
\newcommand{\eimu}{\mbf{e}_{\text{gyro}}}
\newcommand{\yimu}{\by_{\text{gyro}}}
\newcommand{\ts}[1]{t_{#1}}

\begin{document}

\title{Need for Speed: Fast Correspondence-Free Lidar-Inertial Odometry \\ Using Doppler Velocity}

\author{David J. Yoon$^{1}$, Keenan Burnett$^{1}$, Johann Laconte$^{1}$, Yi Chen$^{2}$, Heethesh Vhavle$^{2}$, \\ Soeren Kammel$^{2}$, James Reuther$^{2}$, and Timothy D. Barfoot$^{1}$ 
\thanks{$^{1}$University of Toronto Institute for Aerospace Studies (UTIAS), 4925 Dufferin St, Ontario, Canada.
$^{2}$Aeva Inc., Mountain View, CA 94043, USA.
\texttt{\{david.yoon, keenan.burnett, johann.laconte\}@robotics.utias.utoronto.ca, \{ychen, heethesh, soeren, jreuther\}@aeva.ai, tim.barfoot@utoronto.ca}}
}

\maketitle
\begin{abstract}
  In this paper, we present a fast, lightweight odometry method that uses the Doppler velocity measurements from a Frequency-Modulated Continuous-Wave (FMCW) lidar without data association. FMCW lidar is a recently emerging technology that enables per-return relative radial velocity measurements via the Doppler effect. Since the Doppler measurement model is linear with respect to the 6-degrees-of-freedom (DOF) vehicle velocity, we can formulate a linear continuous-time estimation problem for the velocity and numerically integrate for the 6-DOF pose estimate afterward. The caveat is that angular velocity is not observable with a single FMCW lidar. We address this limitation by also incorporating the angular velocity measurements from a gyroscope. This results in an extremely efficient odometry method that processes lidar frames at an average wall-clock time of 5.64ms on a single thread, well below the 10Hz operating rate of the lidar we tested. We show experimental results on real-world driving sequences and compare against state-of-the-art Iterative Closest Point (ICP)-based odometry methods, presenting a compelling trade-off between accuracy and computation. We also present an algebraic observability study, where we demonstrate in theory that the Doppler measurements from multiple FMCW lidars are capable of observing all 6 degrees of freedom (translational and angular velocity).
\end{abstract}

\section{Introduction}
  \label{sec:introduction}

Lidar sensors have proven to be a reliable modality for vehicle state estimation in a variety of applications such as self-driving, mining, and search \& rescue. Modern lidars are long range, high resolution, and relatively unaffected by lighting conditions. State-of-the-art estimation is achieved by algorithms that geometrically align lidar pointclouds through an iterative process of nearest-neighbour data association (i.e., \ac{ICP}-based methods \cite{besl_pami92, pomerleau_15}). 

\begin{figure}[t]
  \centering
  \includegraphics[width=\linewidth]{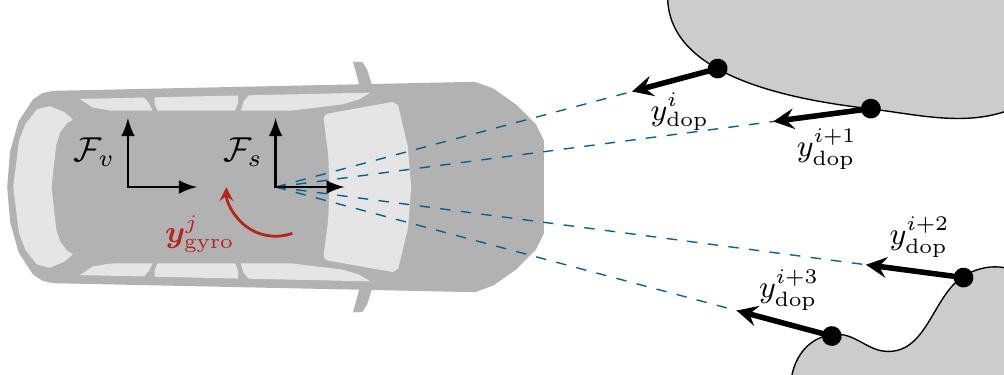}
  \caption{A 2D illustration of our method. We use the Doppler measurements from a \ac{FMCW} lidar and the angular velocity measurements from a gyroscope to efficiently estimate 6-DOF vehicle motion without data association.}
  \label{fig:intro}
  \vspace{-0.2in}
\end{figure}

However, alignment relying on scene geometry can fail in degenerate environments such as tunnels, bridges, or long highways with a barren landscape. \ac{FMCW} lidar is a recent type of lidar sensor that additionally measures per-return relative radial velocities via the Doppler effect (see Figure \ref{fig:intro}). Incorporating these \textit{Doppler measurements} into \ac{ICP}-based methods has recently been demonstrated to substantially improve estimation robustness in these difficult scenarios \cite{hexsel_rss22,yuchen_ral23}.

\ac{ICP}-based methods perform accurately, but are relatively expensive in computation due to the iterative data association. More computationally efficient odometry may be desirable to leave compute available for other processes in an autonomous navigation pipeline (e.g., localization, planning, control). We present an efficient lidar odometry method that leverages the Doppler measurements and does not perform data association as it is a major computational bottleneck.

In this work, we propose a lightweight odometry method that estimates for the 6-\ac{DOF} vehicle velocity, which can afterward be numerically integrated into a $SE(3)$ pose estimate. Velocity, rather than pose, is estimated because the Doppler measurement model is linear with respect to the vehicle velocity, permitting a linear continuous-time estimation formulation. A caveat is that the vehicle velocity is not fully observable with a single \ac{FMCW} lidar. We address this problem by also using the gyroscope measurements from an \ac{IMU}, conveniently built into the Aeries I \ac{FMCW} lidar that we used for testing. The resulting method produces pose estimates at an average wall-clock time of 5.64ms on a single thread, which is substantially lower than the 100ms time budget required for the 10Hz operating rate of the lidar. In comparison to the state of the art, we believe we offer a compelling trade-off between accuracy and performance.

The following are the main contributions of this paper:
\begin{itemize}
  \item A lightweight, linear estimator for the vehicle velocity using Doppler and gyroscope measurements.
  \item An observability study of the vehicle velocity estimated using Doppler measurements, showing in theory that the Doppler measurements from multiple \ac{FMCW} lidars can result in the observability of all 6 degrees of freedom (encouraging future research on this topic).
  \item Experimental results on real-world driving sequences for our proposed odometry method and comparisons to state-of-the-art \ac{ICP}-based methods.
\end{itemize}

The remainder of the paper is as follows: Section \ref{sec:relatedwork} presents relevant literature; Section \ref{sec:methodology} presents the odometry methodology; Section \ref{sec:observability} presents the observability study; Section \ref{sec:experiments} presents the results and analysis; and finally, Section \ref{sec:conclusion} presents the conclusion and future work. 


\section{Related Work}
  \label{sec:relatedwork}

\subsection{ICP-based Lidar Odometry}
\ac{ICP} estimates the relative transformation between two pointclouds by iteratively re-associating point measurements via nearest-neighbour search \cite{besl_pami92, pomerleau_15}. Lidar odometry methods that achieve state-of-the-art performance apply this simple-but-powerful concept of nearest-neighbour data association in a low-dimensional space (e.g., Cartesian). In this paper, we refer to these algorithms as \ac{ICP}-based. LOAM \cite{zhang_ar17}, a top contender in the publicly available KITTI odometry benchmark \cite{geiger_cvpr12}, extracts edge and plane features, and iteratively matches them via nearest-neighbour association. SuMa \cite{behley_rss18} matches measurements using projective data association (i.e., azimuth-elevation space) and leverages GPU computation to perform this operation quickly.


Modern lidars output high-resolution, \ac{3D} pointclouds by mechanical actuation. Consequently, pointclouds acquired from a moving vehicle will be motion distorted, similar to a rolling-shutter effect. One can motion-compensate (de-skew, or undistort) the data as a preprocessing step \cite{ye_icra19,vizzo2023ral}. Alternatively, data can be incorporated at their exact measurement times by estimating a continuous-time trajectory \cite{furgale_icra12,anderson_icra13,barfoot_rss14}. Continuous-time \ac{ICP}-based methods have been successfully demonstrated in several works \cite{wong_ral20,dellenbach_icra22}. State-of-the-art lidar odometry methods address the motion compensation problem and are capable of achieving highly accurate, real-time performance \cite{pan2021mulls,dellenbach_icra22,vizzo2023ral}. Pan et al. \cite{pan2021mulls} extract low-level geometric features to apply multiple error metrics in their \ac{ICP} optimization. Dellenbach et al. \cite{dellenbach_icra22} use a sparse voxel data structure for downsampling and nearest-neighbour search in a single-threaded implementation. Vizzo et al. \cite{vizzo2023ral} demonstrate faster performance with comparable accuracy by proposing a simplified registration pipeline that requires few tuning parameters in a multi-threaded implementation.

Recently, \ac{FMCW} lidars have been demonstrated to be beneficial in improving odometry. Hexsel et al. \cite{hexsel_rss22} incorporate Doppler measurements into \ac{ICP} to improve estimation in difficult, geometrically degenerate locations. Wu et al. \cite{yuchen_ral23} improve upon this by using a continuous-time estimator, not requiring motion compensation as a preprocessing step.

A major bottleneck in lidar odometry is data association due to (i) the vast amount of data, and (ii) the need for iterative data association. With the introduction of Doppler measurements from \ac{FMCW} lidars, we propose a more efficient odometry method that avoids data association entirely.





\subsection{Inertial Measurements and Lidar}
Lidar odometry algorithms have proven to be highly accurate in nominal conditions, but will struggle to perform in geometrically degenerate environments (e.g., long tunnels, barren landscapes). Using \ac{IMU} data is a way of handling these difficult scenarios, with the added benefit of being able to use the \ac{IMU} to motion-compensate pointclouds as a preprocessing step. Loosely coupled methods may only use the \ac{IMU} data for motion compensation \cite{palieri2020locus}, but can also fuse the pose estimates from pointcloud alignment with \ac{IMU} data downstream \cite{tagliabue2021lion}. Zhao et al. \cite{Zhao2021} implement an odometry estimator for each sensor modality, where each estimator uses the outputs of the other estimators as additional observations. Chen et al. \cite{chen2023direct} combine their pose estimates from \ac{ICP} with IMU data using a hierarchical geometric observer. Tightly coupled methods incorporate \ac{IMU} data into the pointcloud alignment optimization directly, which has been shown using an iterated extended Kalman filter \cite{Qin2020,xu2022fast} and factor graph optimization over a sliding window \cite{ye_icra19,shan_iros20}.

Our work differs from existing lidar-inertial methods in both motivation and implementation. We propose an approach that does not require data association by using the Doppler measurements of a \ac{FMCW} lidar. Our motivation for using \ac{IMU} data is to compensate for the degrees of freedom not observable from the Doppler measurements of a single \ac{FMCW} lidar. We only require gyroscope data (i.e., angular velocities) and exclude the accelerometer\footnote{The gravity vector would require a nonlinear estimator for orientation.}, permitting a linear continuous-time formulation. We do not require pre-integration of the gyroscope data \cite{forster2016manifold}, and instead efficiently incorporate data at their exact measurement times.

\subsection{Radar Odometry}
\ac{FMCW} is a relatively new technology for lidar, but not for radar. Radar, in contrast to \ac{FMCW} lidar, returns \ac{2D} detections (azimuth and range). Similar to our proposed method, Kellner et al. \cite{kellner_itsc13} estimate vehicle motion using radar Doppler measurements without data association. Using a single radar, they estimate a 2-\ac{DOF} vehicle velocity (forward velocity and yaw rotation) by applying a kinematic constraint on the lateral velocity to be zero. Using multiple radars allowed them to estimate a 3-\ac{DOF} vehicle velocity \cite{kellner2014instantaneous}. As radars produce \ac{2D} data in lesser quantities compared to lidar, Kellner et al. \cite{kellner_itsc13, kellner2014instantaneous} limited their experiments to driven sequences that were a few hundred meters in length. Our \ac{FMCW} lidar produces thousands of \ac{3D} measurements at a rate of 10Hz. We take advantage of the richer data by efficiently applying them in a continuous-time linear estimator, and demonstrate reasonably accurate odometry over several kilometers.

Kramer et al. \cite{kramer2020radar} estimate the motion of a handheld sensor rig by combining radar Doppler measurements and \ac{IMU} data. Park et al. \cite{park20213d} estimate for 6-\ac{DOF} motion by first estimating the \ac{3D} translational velocities, then loosely coupling them with \ac{IMU} data in a factor graph optimization. We similarly use \ac{IMU} data to help estimate 6-\ac{DOF} motion. However, we only use the gyroscope data to keep the estimator linear and efficient. We also present an observability study, demonstrating in theory how multiple \ac{FMCW} sensors constrain all degrees of freedom of the velocity.

\section{Methodology}
  \label{sec:methodology}
  \subsection{Problem Formulation}
We formulate our odometry as linear continuous-time batch estimation for the 6-\ac{DOF} vehicle body velocities using a \ac{MAP} \cite{barfoot_se17} objective. This is possible due to how the Doppler velocity and gyroscope measurement models are both linear with respect to the vehicle velocity. A continuous-time formulation allows each measurement to be applied at their exact measurement times efficiently. The relative pose estimate can be computed via numerical integration as a final step. 

The proposed method is extremely lightweight as the estimation problem is linear and we do not require data association for the lidar data. We can apply our method online by incrementally marginalizing out all past velocity state variables (e.g., a linear Kalman filter that handles measurements asynchronously (continuous-time)).

\subsection{Motion Prior}
We apply the continuous-time estimation framework of Barfoot et al. \cite{barfoot_rss14} to estimate the trajectory as a \ac{GP}. We model our vehicle velocity prior as \ac{WNOA} \cite{barfoot_se17}, 
\begin{equation}
    \dot{\bv}(t) = \mbf{w}(t), \quad \mbf{w}(t) \sim \mathcal{GP} (\mbf{0}, \mbf{Q}_c \delta(t - t')),
\end{equation}
where $\mbf{w}(t)$ is a (stationary) zero-mean GP with a power spectral density matrix, $\mbf{Q}_c$.


In addition, we found it beneficial to incorporate vehicle kinematics by penalizing velocities in specific dimensions. We center our vehicle frame at the rear-axle of the vehicle and orient it such that the $x$-axis points forward, $y$-axis points left, and $z$-axis points up. We can penalize velocities in the lateral, vertical, roll, and pitch dimensions:
\begin{equation}
  \mbf{e}^\text{kin}_k = \mbf{H} \bv_k,
\end{equation}
where a constant $\mbf{H}$ extracts the dimensions of interest.

\subsection{Measurement Models}
\label{subsec:meas}
We use the same Doppler measurement model as presented by Wu et al. \cite{yuchen_ral23}. The (scalar) linear error model is
\begin{equation}
\edop^i = \ydop^i - 
\frac{1}{({\gq^i_s}^T {\gq^i_s})^{\frac12}} \begin{bmatrix}{\gq^i_s}^T & \bm{0}\end{bmatrix} \!\!\adT_{sv} \bv(\ts{i}) - h(\mbs{\psi}^i), \label{eq:doppler} 
\end{equation}
where $\ydop^i$ is the $i^\text{th}$ Doppler measurement, ${\gq^i_s} \in \mathbb{R}^3$ are the corresponding point coordinates in the sensor frame, $\adT_{sv} \in \text{Ad}\left( SE(3)\right)$ is the (known) extrinsic adjoint transformation between the sensor and vehicle frames, and $\bv(\ts{i}) \in \mathbb{R}^6$ is our continuous-time vehicle velocity queried at the corresponding measurement time.
We are simply projecting the vehicle velocity into the radial direction of the measurement in the sensor frame (see Figure \ref{fig:intro}).

\begin{figure}[t]
  \centering
  \includegraphics[width=\linewidth, trim=0 5 0 0, clip]{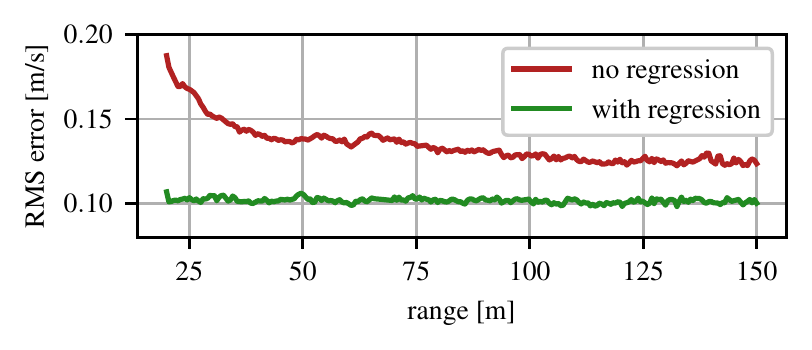}
  \caption{A plot of the root-mean-square Doppler error versus measurement range aggregated over a driven test data sequence. Discretizing the azimuth-elevation domain and learning a linear regression model in each bin with range as the input feature noticeably improves the error. Note that the linear dependance on range is not visible in this plot as we aggregate over all azimuth-elevation bins.}
  \label{fig:range_bias}
  \vspace{-0.2in}
\end{figure}

We identified a non-zero bias in the Doppler measurements through experimentation. Wu et al. \cite{yuchen_ral23} calibrated this bias using stationary lidar data collected from a flat wall at a fixed distance. Measurements were partitioned into (approximately) uniform bins of azimuth and elevation, and a constant bias was calibrated for each bin. In our recent work, we further identified that the bias has an approximately linear dependance on the range measurement. Therefore we instead model the Doppler velocity bias using a linear regression model, $h(\mbs{\psi})$, with input feature vector, $\mbs{\psi} = [1 \quad (\gq^T \gq)^{\frac12}]^T$, for each azimuth-elevation bin. We will investigate other input features (e.g., incidence angle, intensity) and nonlinear regression models (if the need arises) in future work to improve performance. Figure \ref{fig:range_bias} demonstrates a before-and-after comparison of applying our learned regression models on a real-world test sequence\footnote{Evaluated using the velocity estimates from an Applanix POS-LV as groundtruth.}.

Partitioning the lidar data by azimuth and elevation also has the effect of downsampling since we keep one measurement per azimuth-elevation bin. We uniformly partition along the azimuth by $0.2^\circ$. The data are already partitioned in elevation by the scan pattern of the sensor, which produces 80 horizontal sweeps. This downsampling effectively projects each lidar frame into a $80 \times 500$ image. In practice, a raw lidar frame has approximately 100,000 measurements, which is then downsampled to 10,000 to 20,000 depending on the scene geometry\footnote{This is typically an order of magnitude more data than what is used in state-of-the-art \ac{ICP}-based methods \cite{dellenbach_icra22,yuchen_ral23}.}.

In addition to the Doppler measurements, we use gyroscope data to compensate for the degrees of freedom not observed by the Doppler velocities of a single lidar (see Section \ref{sec:observability} for an algebraic study). The error model is
\begin{equation}
\eimu^j = \yimu^j - \mbf{R}_{sv} \Dw \bv(\ts{j}),
\end{equation}
where $\yimu^j \in \mathbb{R}^3$ is the $j^\text{th}$ angular velocity measurement in the sensor frame, $\mbf{R}_{sv} \in SO(3)$ is the known extrinsic rotation between the sensor frame and vehicle frame, and $\bv(\ts{j})$ is our continuous-time velocity of the vehicle frame queried at the corresponding measurement time. The constant $3 \times 6$ projection matrix $\Dw$ removes the translational elements of the body velocity, leaving only the angular elements. Similar to the Doppler model, this function is linear with respect to the vehicle velocity. We verified empirically that the gyroscope bias is reasonably constant. We apply an offline calibration for a constant bias on training data. In future work, we plan on improving this aspect of the implementation by including the bias as part of the state.

\subsection{Estimation}
The measurement factors of our estimation problem are
\begin{equation}
  \phi^{i}_{\text{dop}} = \frac{1}{2} (\edop^i)^2 \, {R^{-1}_\text{dop}}
\end{equation} 
for the Doppler measurements and
\begin{equation}
    \phi^{j}_{\text{gyro}} = \frac{1}{2} {\eimu^j}^T \mbf{R}^{-1}_{\text{gyro}} \eimu^j
\end{equation}
for the gyroscope measurements, where $R_{\text{dop}}$ is the Doppler measurement variance and $\mbf{R}_{\text{gyro}}$ is the gyroscope covariance.

The motion prior factor of our \ac{WNOA} prior is
\begin{equation}
    \phi^k_{\text{wnoa}} = \frac{1}{2} (\bv_k - \bv_{k-1})^T \mbf{Q}^{-1}_k (\bv_k - \bv_{k-1})
\end{equation}
for the set of discrete states, $\bv_k$, that we estimate in our continuous-time trajectory. We space these discrete states uniformly in time, corresponding to the start and end times of each lidar frame, making $\mbf{Q}_k = (\ts{k} - \ts{k-1}) \mbf{Q}_c$ constant. This prior conveniently results in linear interpolation in time for our velocity-only state \cite{barfoot_se17}:
\begin{equation}
\bv(\tau) = (1-\alpha) \bv_{k} + \alpha \bv_{k+1}, \quad \alpha = \frac{\tau - \ts{k}}{\ts{k+1} - \ts{k}} \in [0, 1],
\end{equation}
with $\tau \in [\ts{k}, \ts{k+1}]$. The vehicle kinematics factor is
\begin{equation}
  \phi^k_{\text{kin}} = \frac{1}{2} (\mbf{H} \bv_k)^T \mbf{Q}_z^{-1}  (\mbf{H} \bv_k),
\end{equation}
where $\mbf{Q}_z$ is the corresponding covariance matrix\footnote{$\mbf{R}_\text{gyro}$, $\mbf{Q}_c$, and $\mbf{Q}_z$ were empirically tuned as diagonal matrices. All noise parameter values will be accessible from our implementation.}.

Our \ac{MAP} objective function is
\begin{equation}
    J = \sum_k (\phi^k_{\text{wnoa}} + \phi^k_{\text{kin}}) + \sum_i \phi^{i}_{\text{dop}} + \sum_j \phi^{j}_{\text{gyro}}.
\end{equation}
Differentiating this objective with respect to the state and setting it to zero for an optimum will result in a linear system $\mbs{\Sigma}^{-1} \bv^* = \mbf{b}$, where $\bv = [\bv_1^T \quad \bv_2^T \quad \dots \quad \bv_K^T]^T$ is a stacked vector of all the vehicle velocities, $\mbs{\Sigma}^{-1}$ is the corresponding block-tridiagonal inverse covariance, and $\bv^*$ can be computed using a sparse solver.

\begin{figure}[t]
  \centering
  \includegraphics[width=\linewidth]{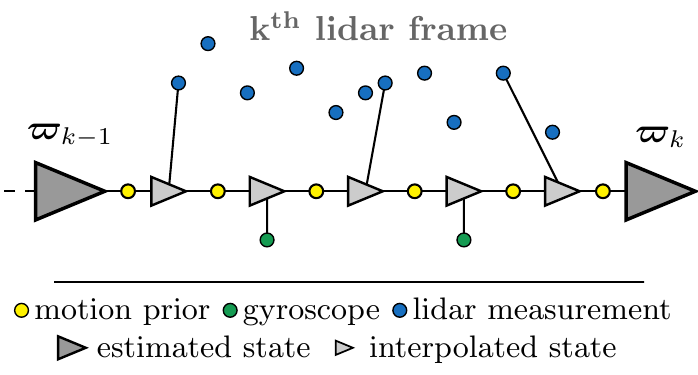}
  \caption{An illustration of the factors involved in the online velocity estimation problem. The Doppler and gyroscope measurements are applied at their exact measurement times using our continuous-time interpolation scheme. There is no data association for the lidar measurements. We marginalize out past state variables (i.e., $\mbs{\varpi}_{k-1}$), resulting in a filter for the latest velocity, $\mbs{\varpi}_k$.}
  \label{fig:factorgraph}
  \vspace{-0.2in}
\end{figure}

Figure \ref{fig:factorgraph} illustrates the states and factors in our online problem. For the latest lidar frame, $k$, the Doppler measurements are incorporated at their measurement times using our continuous-time interpolation scheme. The gyroscope measurements are similarly handled at their respective measurement times. We incrementally marginalize\footnote{In the interest of space, see Barfoot \cite{barfoot_se17} for the details.} out older state variables, $\mbs{\varpi}_i$, where $i < k$, and estimate the latest velocity, $\mbs{\varpi}_k$ (i.e., a filter implementation).

After the vehicle velocity is estimated, we approximate the relative pose estimate by numerically sampling $\bv(t)$ with a small timestep, $\triangle t$, and creating a chain of $SE(3)$ matrices spanning the time interval \cite{anderson_icra13}:
\begin{align}
  &\mbf{T}_{k, k-1} \approx  \exp(\triangle t \, \bv(\ts{k})^\wedge) \dots \\
    &\quad \times \exp(\triangle t \, \bv(\ts{k-1} + 2 \triangle t)^\wedge) \exp(\triangle t \, \bv(\ts{k-1} + \triangle t)^\wedge), \nonumber
\end{align}
where $\exp(\cdot)$ is the exponential map\footnote{The $(\cdot)^\wedge$ here is for $\mathbb{R}^6$ and is an overloading of the $(\cdot)^\wedge$ for $\mathbb{R}^3$. See Barfoot \cite{barfoot_se17} for the details.}, $\bv(t)$ is the vehicle velocity interpolated between boundary velocities $\bv_{k-1}$ and $\bv_k$, and we have divided the time interval by $S$ steps, making $\triangle t = (\ts{k} - \ts{k-1})/S$. In practice, our vehicle pose estimate drifts while the vehicle is stationary (e.g., no movement due to traffic). This is easily detected by checking a tolerance (we use 0.03m/s) on the forward translational speed estimate. If the speed of a boundary estimate is less than the tolerance, we set that boundary estimate $\bv_i = \mbf{0}$ before interpolation to mitigate pose drift.

\subsection{RANSAC for Outlier Rejection} 
Outliers in the Doppler measurements are often caused by erroneous reflections and moving objects in the environment. Fortunately, each lidar frame is dense and, in practice, the majority of the measurements are of the stationary environment (inliers). Similar to Kellner et al. \cite{kellner_itsc13}, we found RANSAC \cite{ransac} to be a suitable method for outlier filtering. 

We classify between inliers and outliers using a constant threshold (0.2m/s) on the Doppler error model (\ref{eq:doppler}). We run RANSAC on each lidar frame independently. We assume the vehicle velocity is constant throughout each frame and solve for it using two randomly sampled Doppler measurements. The solve is made observable by enforcing vehicle kinematic constraints, i.e., we solve for a 2-\ac{DOF} velocity\footnote{Our vehicle frame is oriented such that the $x$-axis points forward, $y$-axis points left, and $z$-axis points up.} $\bv = [v \quad 0 \quad 0 \quad 0 \quad 0 \quad \omega]$. We could include the gyroscope measurements and solve for more dimensions, but found the benefits in performance to be minor. In practice, 20 iterations of RANSAC was sufficient for each lidar frame.

\section{Observability Study}
\label{sec:observability}
\subsection{Observability Study - Multiple FMCW Lidars}
  We present an observability study for the 6-\ac{DOF} vehicle velocity using Doppler measurements from multiple \ac{FMCW} lidars. In order to simplify the proof, we focus on estimating the vehicle velocity over the interval of one lidar frame, assuming that the data from multiple lidars are synchronized and each have $m$ measurements. We also remove the continuous-time aspect of the problem by assuming the vehicle velocity is constant throughout the frame duration.
  
  For the $i^\text{th}$ measurement seen by the $j^\text{th}$ sensor,
  \begin{equation}
    \begin{aligned}
      \edop^{ij} 
                   &= \ydop^{ij} - \frac{1}{({\gqi}^T\gqi)^\frac12}\begin{bmatrix}{\gqi}^T & \bm{0}\end{bmatrix} \gT{jv}\gvp{}{}\\
                   &= \ydop^{ij} - \frac{1}{({\gqi}^T\gqi)^\frac12} \begin{bmatrix}{\gqi}^T\gR{jv} & {\gqi}^T \gr{j}{vj}^\wedge \gR{jv}\end{bmatrix} \gvp{}{} \\
                 &= \ydop^{ij} - \bm{c}_{ij}^T \gvp{}{},
    \end{aligned}
  \end{equation}
  where the additional superscript $j$ indicates the sensor\footnote{$\mbf{q}^{ij}_{f}$ are the coordinates of the $i^\text{th}$ point from sensor $j^\text{th}$, in frame $f$.}, $\adT_{sv} \in \text{Ad}\left( SE(3)\right)$ is the extrinsic adjoint transformation between the $j^\text{th}$ sensor and vehicle frames, and
  \begin{equation}
    \bm{c}_{ij}^T = \begin{bmatrix} \gqh{v}{ij}^T & \gqh{v}{ij}^T\gr{v}{vj}^\wedge \end{bmatrix}, \quad \gqh{v}{ij}=\frac{\gR{jv}^T\gqi}{\norm{\gqi}}.
  \end{equation}
  Note how the measurement model does not depend on the magnitude (range) of $\gq$ since $\gqh{}{}$ are unit vectors.
  We define the stacked quantity $\gC{j} = [\bm{c}_{1j} \cdots \bm{c}_{mj}]^T$ for sensor $j$. In the case of $N$ sensors, we have
  \begin{equation}
    \begin{aligned}
    \gC{}^T\gC{} &= \sum_{j=1}^N \gC{j}^T\gC{j} 
                 = \sum_j \begin{bmatrix} 
                            \bm{Q}_j & \bm{Q}_j\gr{v}{vj}^\wedge \\
                            {\gr{v}{vj}^\wedge}^T\bm{Q}_j & {\gr{v}{vj}^\wedge}^T\bm{Q}_j\gr{v}{vj}^\wedge
                           \end{bmatrix}
    \end{aligned},
  \end{equation}
  where $\bm{Q}_j = \sum_i\gqh{v}{ij}\gqh{v}{ij}^T$ is the sum of the outer product of the points seen by the $j^\text{th}$ sensor. 
  The velocity is fully observable from a single lidar frame if and only if $\gC{}^T\gC{}$ is full rank, or equivalently that the nullspace of $\gC{}^T\gC{}$ has dimension zero \cite{barfoot_se17}.
  In the following, we assume $\bm{Q}_j$ to be full rank (best case scenario), meaning that the unit velocities seen by the $j^\text{th}$ sensor are not all contained in a line or a plane. In the case of a 3D lidar sensor, $\bm{Q}_j$ will always be full rank regardless of the environment geometry.
  \begin{lemma}
  \label{lemma:nullspaceSum}
    Let $\bA$ and $\bB$ be two symmetric positive semidefinite matrices. 
    Then, we have
    $$ \nullspace{\bA+\bB} = \nullspace \bA \cap \nullspace \bB. $$
  \end{lemma}
  A proof of this lemma is given in \refappendix{app:proof}.

  First, note that each member can be factorized as
  \begin{equation}
    \gC{j}^T\gC{j} = 
          \underbrace{
            \begin{bmatrix} \I & \bm{0} \\ 
              {\gr{v}{vj}^\wedge}^T & \I \\
          \end{bmatrix}}_\text{full rank}
          \underbrace{
            \begin{bmatrix} \bm{Q}_j & \bm{0}\\ 
              \bm{0} & \bm{0}\\
          \end{bmatrix}}_\text{PSD, rank$=\!3$}
          \underbrace{
            \begin{bmatrix} \I & \gr{v}{vj}^\wedge \\ 
              \bm{0} & \I \\
          \end{bmatrix}}_\text{full rank},
  \end{equation}
  thus being positive semidefinite. Using \autoref{lemma:nullspaceSum}, we find the nullspace of $\gC{}^T\gC{}$ using the nullspace of each member of the sum. The nullspace of $\gC{j}^T\gC{j}$ is
  \begin{equation}
    \begin{aligned}
      \nullspace{\gC{j}^T\gC{j}} &= \nullspace{
        \begin{bmatrix} \bm{Q}_j & \bm{0}\\ 
              \bm{0} & \bm{0}\\
          \end{bmatrix}
            \begin{bmatrix} \I & \gr{v}{vj}^\wedge \\ 
              \bm{0} & \I \\
          \end{bmatrix}
    } \\
                                 &= \left\{
            \begin{bmatrix} \I & \gr{v}{vj}^\wedge \\ 
              \bm{0} & \I \\
            \end{bmatrix}^{-1}
            \begin{bmatrix} \bm{0} \\ \bm{k} \end{bmatrix},
            \bm{k} \in \mathbb{R}^3
                                 \right\} \\
                                 &= \left\{
                                   \begin{bmatrix} -\gr{v}{vj}^\wedge\bm{k} \\ \bm{k} \end{bmatrix},
            \bm{k} \in \mathbb{R}^3
                                 \right\}.
    \end{aligned}
  \end{equation}
  Thus for one sensor, $\gC{}^T\gC{}$ is rank deficient by 3.
  For two sensors, the nullspace of $\gC{1}^T\gC{1} + \gC{2}^T\gC{2}$ is
  \begin{equation}
    \begin{aligned}
                                &\left\{
                                   \begin{bmatrix} -\gr{v}{v1}^\wedge\bm{k} \\ \bm{k} \end{bmatrix}, \bm{k} \in \mathbb{R}^3
                                 \right\} \cap
                                 \left\{
                                   \begin{bmatrix} -\gr{v}{v2}^\wedge\bm{l} \\ \bm{l} \end{bmatrix}, \bm{l} \in \mathbb{R}^3
                                 \right\} \\
                                                  &= 
                              \begin{cases}
                                 \left\{
                                   \begin{bmatrix} \alpha\gr{v}{v2}^\wedge\gr{v}{v1} \\ \alpha(\gr{v}{v2}-\gr{v}{v1}) \end{bmatrix}, \alpha \in \mathbb{R} 
                                 \right\} & \text{if }\gr{v}{v1}\not=\gr{v}{v2} \\[1em]
                                 \left\{
                                 \begin{bmatrix} -\gr{v}{v1}^\wedge\bm{k} \\ \bm{k} \end{bmatrix}, \bm{k} \in \mathbb{R}^3 
                                  \right\}& \text{if }\gr{v}{v1}=\gr{v}{v2},
                              \end{cases}
    \end{aligned}
  \end{equation}
  therefore being of dimension 1 if the two sensors are not at the same position, and dimension 3 otherwise.
  Adding a third sensor, we obtain
  \begin{equation}
    \begin{aligned}
                                 &\left\{
                                   \begin{bmatrix} \alpha\gr{v}{v2}^\wedge\gr{v}{v1} \\ \alpha(\gr{v}{v2}-\gr{v}{v1}) \end{bmatrix}, \alpha \in \mathbb{R} 
                               \right\} \cap
                                 \left\{
                                   \begin{bmatrix} -\gr{v}{v3}^\wedge\bm{k} \\ \bm{k} \end{bmatrix}, \bm{k} \in \mathbb{R}^3
                                 \right\} =\\
                                  & 
                                 \left\{
                                   \begin{bmatrix} \alpha\gr{v}{v2}^\wedge\gr{v}{v1} \\ \alpha(\gr{v}{v2}-\gr{v}{v1}) \end{bmatrix}\middle| \gr{v}{v2}^\wedge\gr{v}{v1} \!=\! -\gr{v}{v3}^\wedge\!\left(\gr{v}{v2} - \gr{v}{v1}\right)\!, \alpha \in \mathbb{R}\!%
                               \right\}.
    \end{aligned}
  \end{equation}
  Looking at the condition, we remark that $\gr{v}{v2}^\wedge\gr{v}{v1}, -\gr{v}{v3}^\wedge\left(\gr{v}{v2} - \gr{v}{v1}\right) \in \gr{v}{v2}^\bot\cap\gr{v}{v3}^\bot$ is necessary, where $(\cdot)^\bot$ denotes the orthogonal complement.
  As such, we can re-write the condition as
  \begin{equation}
    \begin{aligned}
      &%
      \begin{cases}
        \gr{v}{v2}^\wedge\gr{v}{v1} &= \beta \gr{v}{v2}^\wedge\gr{v}{v3}, \quad\beta\in\mathbb{R} \\
        -\gr{v}{v3}^\wedge\left(\gr{v}{v2} - \gr{v}{v1}\right) &= \beta \gr{v}{v2}^\wedge\gr{v}{v3}
    \end{cases} \\
      \Leftrightarrow&%
      \begin{cases}
        \gr{v}{v1} &= \gamma \gr{v}{v2} + \beta\gr{v}{v3}, \quad\gamma\in\mathbb{R} \\
        \beta &= 1 - \gamma
    \end{cases} \\
    \Leftrightarrow&\,\, 
      \gr{v}{v1} = \gamma \gr{v}{v2} + (1-\gamma)\gr{v}{v3}. \\
    \end{aligned}
  \end{equation}
  The nullspace of $\gC{}^T\gC{}$ for three sensors has dimension 1 if all three $\gr{v}{vj}$ are on the same line, and dimension 0 otherwise.
  As such, the full state is observable as long as the three lidar sensors form the vertices of a triangle.
  Note that using induction with \autoref{lemma:nullspaceSum}, we can intuitively add more sensors and the system remains fully observable.

\subsection{Observability Study - Single FMCW Lidar + Gyroscope}
  Considering now the case of one lidar with one gyroscope measurement, we show that adding the gyroscope measurement leads to a fully observable system.
  With one gyroscope measurement, the measurement matrix becomes
  \begin{equation}
        \gC{}'^T = 
        \begin{bmatrix} \gqh{v}{1} & \cdots & \gqh{v}{N} & \bm{0} \\
          {\gr{v}{vj}^\wedge}^T\gqh{v}{1} & \cdots & {\gr{v}{vj}^\wedge}^T\gqh{v}{N} & \gR{sv}
        \end{bmatrix},
  \end{equation}
  leading to
  \begin{equation}
          \gC{}'^T\gC{}' = 
          \underbrace{
            \begin{bmatrix} \I & \bm{0} \\ 
              {\gr{v}{vj}^\wedge}^T & \I \\
          \end{bmatrix}}_\text{full rank}
          \underbrace{
            \begin{bmatrix} \bm{Q} & \bm{0}\\ 
              \bm{0} & \I\\
          \end{bmatrix}}_\text{full rank}
          \underbrace{
            \begin{bmatrix} \I & \gr{v}{vj}^\wedge \\ 
              \bm{0} & \I \\
          \end{bmatrix}}_\text{full rank}.
  \end{equation}
  As such, the system becomes fully observable with only one lidar sensor and one gyroscope.

\section{Experiments}
  \label{sec:experiments}
  \subsection{Experiment Setup}
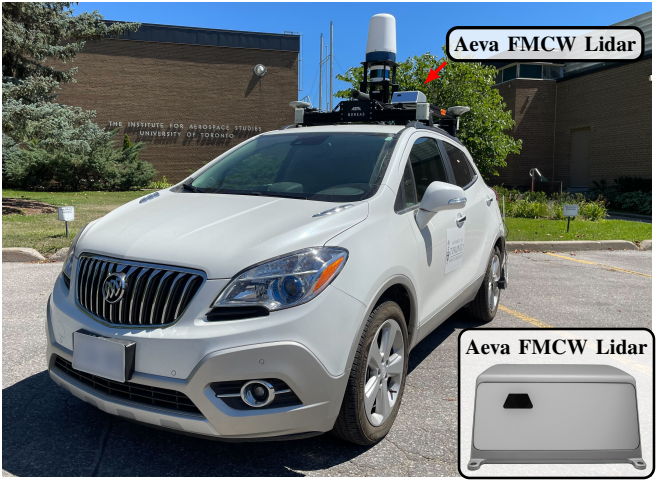
\begin{figure}[t]
  \centering
  \begin{tikzpicture} [
      arrow/.style={>=latex,red, line width=1.25pt},
      block/.style={rectangle, draw, fill=white, fill opacity=0.8, minimum width=4em, text centered, rounded corners, minimum height=1.25em, line width=1.25pt, inner sep=2.5pt}]
    \node[inner sep=0pt] (boreas) {\includegraphics[trim=300 100 0 200, clip, width=\columnwidth]{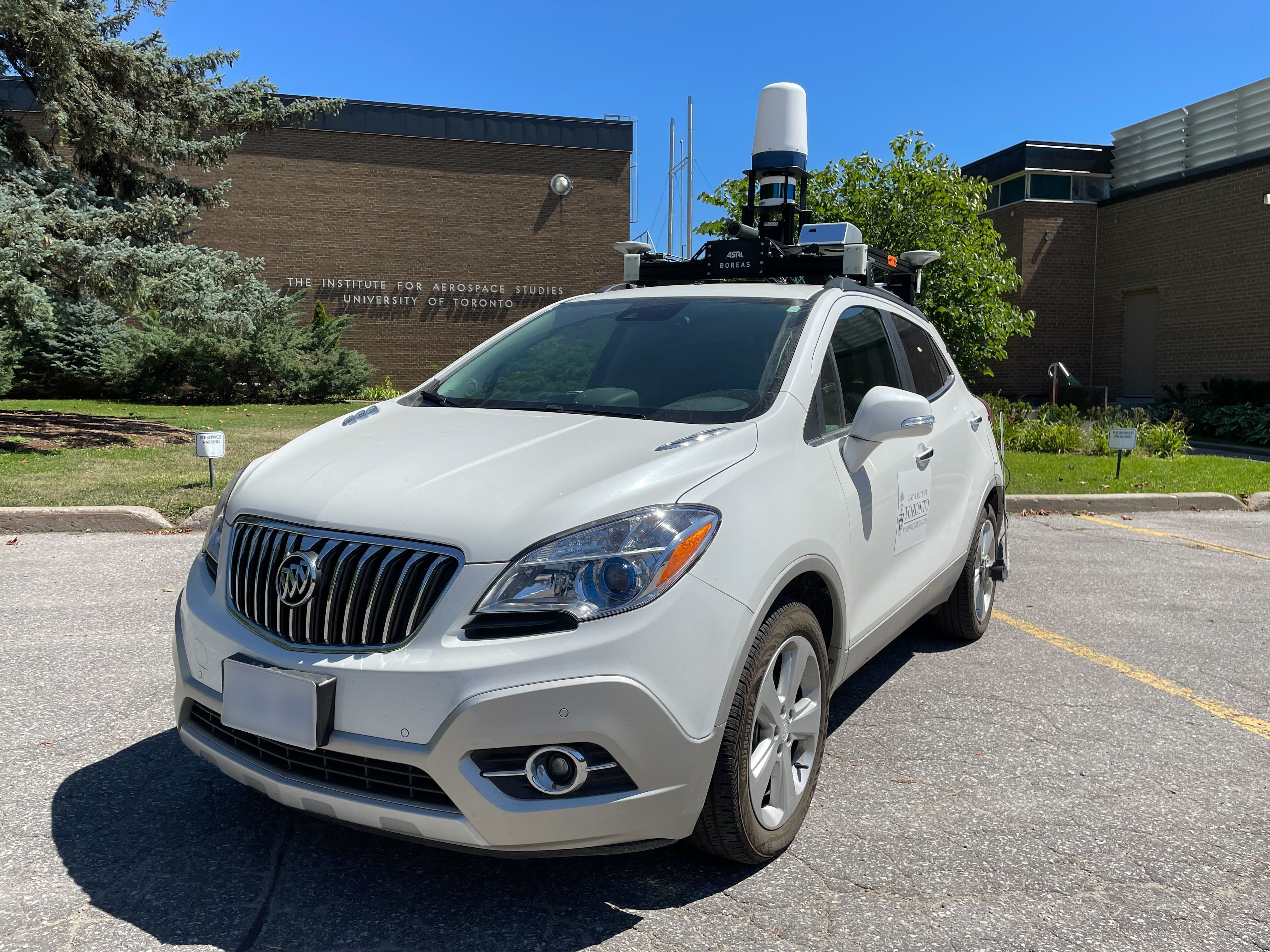}};
    \node (A) at (boreas.center) {};
    \def \L{0.75};

    \node [block, text opacity=1.0] (lidar) at ($ (A) + (29mm, 26mm) $) {\footnotesize\textbf{Aeva FMCW Lidar}};
    \draw[->, arrow] ($ (lidar.south west) + (0mm, 0mm) $) -- ($ (lidar.south west) + (-3mm, -3mm) $) {};

    \node[inner sep=0pt, draw=black, rounded corners, line width=1pt, fill=white, fill opacity=0.8, text opacity=1.0] (aeva) at ($(boreas.center) + (30.5mm, -21.8mm)$) {\includegraphics[trim=400 300 200 0, clip, width=0.3\columnwidth]{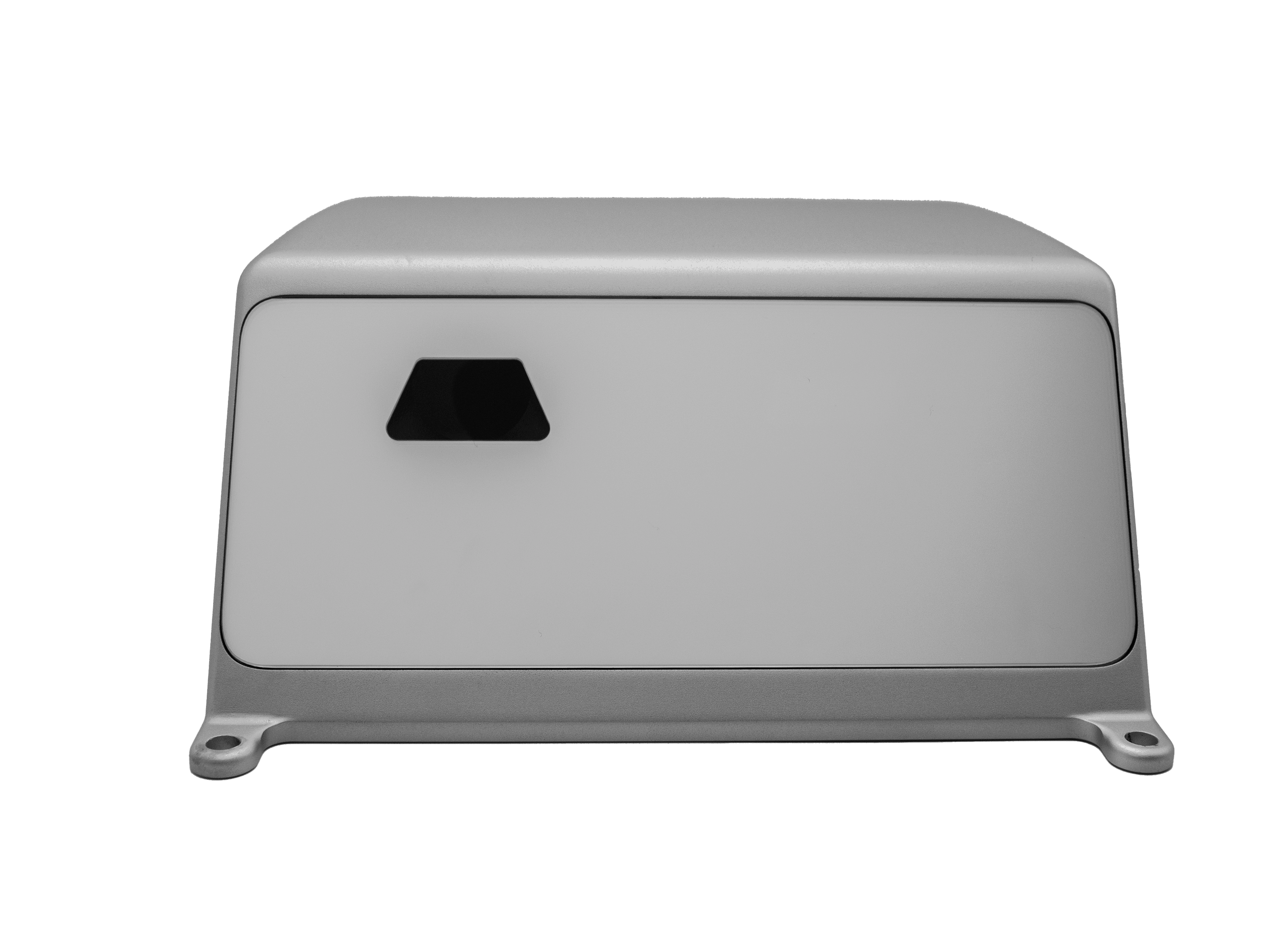}};

    \node [rectangle, text centered, inner sep=0pt] (lidar text) at ($ (aeva.center) + (0mm, 7.5mm) $) {\footnotesize\textbf{Aeva FMCW Lidar}};

  \end{tikzpicture}
  \caption{Our data collection platform, \textit{Boreas}, was established for the \href{https://www.boreas.utias.utoronto.ca}{Boreas dataset} \cite{burnett_ijrr23}. It was recently equipped with an Aeva Aeries I \ac{FMCW} lidar by Wu et al. \cite{yuchen_ral23}.}
  \label{fig:buick}
  \vspace{-0.2in}
\end{figure}

An image of our data collection vehicle, \textit{Boreas}, is shown in Figure \ref{fig:buick}. The vehicle was previously used for the \href{https://www.boreas.utias.utoronto.ca}{Boreas dataset} \cite{burnett_ijrr23} and demonstrating STEAM-ICP with Doppler velocity measurements \cite{yuchen_ral23}. Boreas is equipped with an Aeva Aeries I \ac{FMCW} lidar sensor, which has a horizontal field-of-view of $120^\circ$, a vertical field-of-view of $30^\circ$, a 300m maximum operating range, and a sampling rate of 10Hz. The lidar includes a Bosch BM160 \ac{IMU}, which we use for our experiments. We use the post-processed estimates from an Applanix POS LV as our groundtruth.

We collected 5 data sequences near the University of Toronto Institute for Aerospace Studies. Sequences 1 to 4 follow the \textit{Glen Shields} route of the Boreas dataset \cite{burnett_ijrr23}. Sequence 5 is a different route collected in the same area.

Following existing work, we evaluate odometry using the KITTI error metric, which averages errors over path lengths that vary from 100m to 800m in 100m increments. We present results for two variants of our method: an offline batch implementation and an online filter implementation.

\subsection{Implementation}
We ran all experiments using the same compute hardware\footnote{Lenovo Thinkpad P53, Intel Core i7-9750H CPU.}. Our C++ filter implementation\footnote{Code for the C++ implementation: \url{https://github.com/utiasASRL/doppler_odom}.} is single-threaded. Average wall-clock times of the main modules in our pipeline are:
\subsubsection{Preprocessing (4.19ms)} 
Downsampling by azimuth and elevation (most expensive computation) and evaluation of the linear regression model (negligible computation).
\subsubsection{RANSAC (0.95ms)}
RANSAC for the latest lidar frame to filter out outliers.
\subsubsection{Solve (0.49ms)}
Solving for the latest velocity estimate corresponding to the last timestamp of the latest lidar frame.
\subsubsection{Numerical integration (0.01ms)}
Approximate the latest pose estimate using numerical integration (100 steps).

The total average wall-clock time for processing the latest lidar frame is 5.64ms. This is well under 100ms, which is the requirement for processing Aeva Aeries I lidar data in real time. We note that the Aeries I firmware does not support outputting the raw azimuths, elevations, and range of the lidar points, requiring re-calculation of these values from the coordinates. These calculations result in an additional 1.5ms on average in the preprocessing step, which we have included in the above timing results. Outputting the raw azimuths and elevations is supported in later sensor models (i.e., Aeries II), which will further reduce our compute time.

\begin{figure}[t]
  \centering
  \includegraphics[width=\linewidth, trim=0 5 0 0, clip]{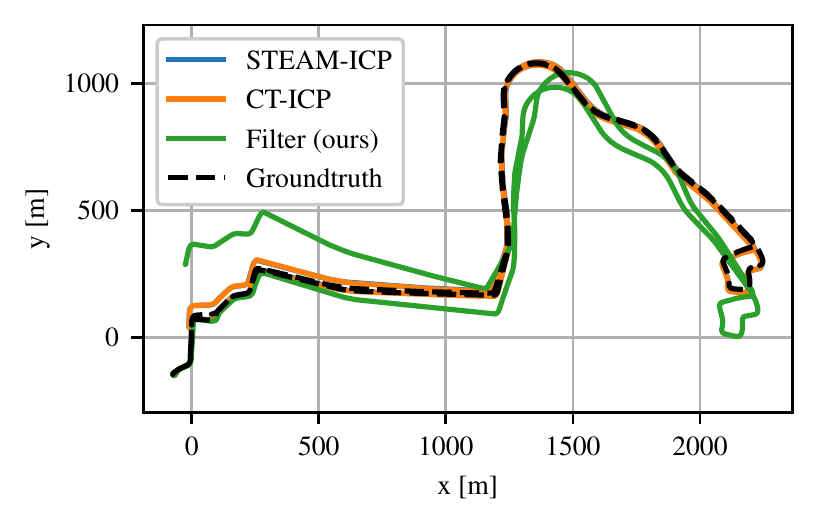}
  \caption{A qualitative plot of the estimated odometry paths on sequence 1 of our collected dataset. Our proposed filter runs at 5.64ms on average for each lidar frame on a single thread. See Figure \ref{fig:path} for an example of sequence 5.}
  \label{fig:path2}
  \vspace{-0.2in}
\end{figure}

\subsection{Results}
We compare our proposed methods (filter and batch) to the current state of the art for \ac{ICP}-based lidar odometry and demonstrate the trade-offs between computation and accuracy. We present results using STEAM-ICP \cite{yuchen_ral23} as a representative of an accurate method that is not optimized for real-time application, while CT-ICP \cite{dellenbach_icra22} is a real-time capable method that is still very accurate. 


\begin{table*}[ht]
  \centering
  \caption{Quantitative results using the KITTI odometry error metrics on our data sequences. For our methods, we train on one sequence and test on the others. All folds are presented with the training sequence in grey (not counted in the average). The average wall-clock time (single-threaded) per lidar frame is shown in brackets for each online method. Multi-threading reduces the runtimes to 201.69ms and 33.91ms for STEAM-ICP and CT-ICP, respectively. Best results are in bold font.}
  \label{tab:exp}

  \begin{tabular}{ l ? c ? c c c c c c ? c c c c c c }
    \toprule
    \midrule
      & Compute [ms] & \multicolumn{6}{c|}{Translation Error [\%]} 
      & \multicolumn{6}{c}{Rotation Error [$^\circ$/(100m)]} \\
    \midrule
      \textbf{ICP-based} & & 01 & 02 & 03 & 04 & 05 & {\textbf{AVG}} & 01 & 02 & 03 & 04 & 05 & {\textbf{AVG}} \\
    \midrule
      STEAM-ICP \cite{yuchen_ral23} & 678.74 & \textbf{0.24} & \textbf{0.24} & \textbf{0.23} & \textbf{0.21} & \textbf{0.25} & \color{blue}\textbf{0.23}
                  & \textbf{0.087} & \textbf{0.083} & \textbf{0.083} & \textbf{0.080} & \textbf{0.084} & \color{blue}\textbf{0.083} \\
      CT-ICP \cite{dellenbach_icra22} & 93.78 & 0.28 & 0.39 & 0.26 & 0.25 & 0.30 & \color{blue}0.29
                 & 0.099 & 0.104 & 0.098 & 0.094 & 0.100 & \color{blue}0.099 \\
    \midrule
      \textbf{Train on 01} (Ours) & & \color{gray}01 & 02 & 03 & 04 & 05 & {\textbf{AVG}} & \color{gray}01 & 02 & 03 & 04 & 05 & {\textbf{AVG}} \\
    \midrule
      Batch & -- & \color{gray}0.94 & 1.20 & 1.22 & 1.03 & 1.03 & \color{blue}1.12
                  & \color{gray}0.317 & 0.403 & 0.392 & 0.356 & 0.376 & \color{blue}0.382 \\
      Filter & \textbf{5.64} & \color{gray}1.10 & 1.26 & 1.22 & 1.16 & 1.06 & \color{blue}1.17
                 & \color{gray}0.386 & 0.463 & 0.440 & 0.415 & 0.397 & \color{blue}0.429 \\       
    \midrule
      \textbf{Train on 02} (Ours) & & 01 & \color{gray}02 & 03 & 04 & 05 & {\textbf{AVG}} & 01 & \color{gray}02 & 03 & 04 & 05 & {\textbf{AVG}} \\
    \midrule
      Batch & -- & 1.20 & \color{gray}1.00 & 1.02 & 0.96 & 0.94 & \color{blue}1.03
                  & 0.387 & \color{gray}0.344 & 0.357 & 0.340 & 0.344 & \color{blue}0.357 \\
      Filter & \textbf{5.64} & 1.34 & \color{gray}1.10 & 1.18 & 1.12 & 1.08 & \color{blue}1.18
                 & 0.448 & \color{gray}0.410 & 0.439 & 0.407 & 0.391 & \color{blue}0.421 \\             
    \midrule
      \textbf{Train on 03} (Ours) & & 01 & 02 & \color{gray}03 & 04 & 05 & {\textbf{AVG}} & 01 & 02 & \color{gray}03 & 04 & 05 & {\textbf{AVG}} \\
    \midrule
      Batch & -- & 1.11 & 1.09 & \color{gray}0.97 & 0.91 & 0.85 & \color{blue}0.99
                  & 0.368 & 0.380 & \color{gray}0.335 & 0.324 & 0.325 & \color{blue}0.349 \\
      Filter & \textbf{5.64} & 1.22 & 1.21 & \color{gray}1.12 & 1.12 & 1.06 & \color{blue}1.15
                 & 0.414 & 0.450 & \color{gray}0.415 & 0.403 & 0.383 & \color{blue}0.412 \\             
    \midrule
      \textbf{Train on 04} (Ours) & & 01 & 02 & 03 & \color{gray}04 & 05 & {\textbf{AVG}} & 01 & 02 & 03 & \color{gray}04 & 05 & {\textbf{AVG}} \\
    \midrule
      Batch & -- & 1.07 & 1.06 & 0.98 & \color{gray}0.89 & 0.85 & \color{blue}0.99
                  & 0.354 & 0.369 & 0.340 & \color{gray}0.323 & 0.328 & \color{blue}0.348 \\
      Filter & \textbf{5.64} & 1.19 & 1.18 & 1.12 & \color{gray}1.10 & 1.03 & \color{blue}1.13
                 & 0.408 & 0.442 & 0.419 & \color{gray}0.401 & 0.383 & \color{blue}0.413 \\             
    \midrule
      \textbf{Train on 05} (Ours) & & 01 & 02 & 03 & 04 & \color{gray}05 & {\textbf{AVG}} & 01 & 02 & 03 & 04 & \color{gray}05 & {\textbf{AVG}} \\
    \midrule
      Batch & -- & 1.03 & 1.03 & 1.04 & 0.92 & \color{gray}0.89 & \color{blue}1.01
                  & 0.346 & 0.354 & 0.347 & 0.326 & \color{gray}0.332 & \color{blue}0.343 \\
      Filter & \textbf{5.64} & 1.17 & 1.14 & 1.15 & 1.09 & \color{gray}1.00 & \color{blue}1.14
                 & 0.408 & 0.425 & 0.424 & 0.398 & \color{gray}0.378 & \color{blue}0.414 \\             
    \midrule
    \bottomrule

  \end{tabular}
  
  \vspace*{-0.2in}
\end{table*}

\begin{figure*}[ht]
  \centering
  \includegraphics[trim={0.1in 0.1in 0.1in 0.0in},clip]{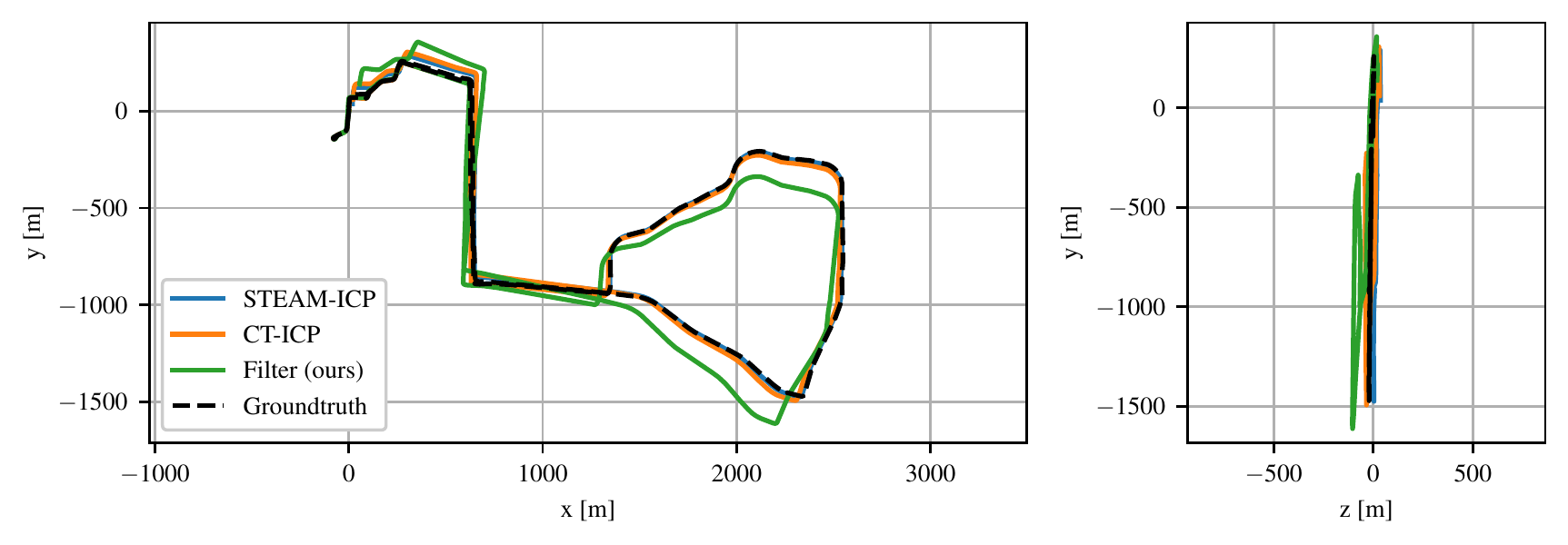}

  \caption{A qualitative plot of the estimated odometry paths on sequence 5. The \ac{ICP}-based methods (STEAM-ICP and CT-ICP) perform extremely well with little drift over several kilometers, but require more compute. Our proposed filter presents a trade-off in accuracy for less computation, executing at 5.64ms on average per lidar frame on a single thread.}
  \label{fig:path}
  \vspace*{-0.2in}
\end{figure*}

As our methods require training for the linear regression models and constant gyroscope bias, we train on one sequence and show test results on the remaining four. Table \ref{tab:exp} shows the results for all five possible folds and compares them to STEAM-ICP and CT-ICP. STEAM-ICP performs better than our filter by a factor of ${\sim}5$ in translation, but runs slower by a factor of ${\sim}120$ on a single thread. A multi-threaded implementation of STEAM-ICP runs slower than our filter (single-threaded) by a factor of ${\sim}36$. CT-ICP performs better than our filter by a factor of ${\sim}4$ in translation, but runs slower by a factor of ${\sim}17$ on a single thread. A multi-threaded implementation of CT-ICP runs slower than our filter (single-threaded) by a factor of ${\sim}6$. Figure \ref{fig:path2} and \ref{fig:path} show a qualitative plot of the estimated paths on sequences 1 and 5, respectively. Our method operates on approximately 10,000 to 20,000 measurements for each lidar frame, while STEAM-ICP and CT-ICP operate on approximately 2,000 to 4,000 measurements.

We believe our method presents a compelling trade-off between accuracy and computational cost. As seen in Figure \ref{fig:path}, our filter performs reasonably over several kilometers such that a loop-closure algorithm could detect points of intersection. This poses a potential application where our lightweight odometry operates at a fast rate, while a slower \ac{ICP}-based optimization can execute in parallel at a slower rate for localization and/or loop closure. As an additional benefit, our odometry can motion-compensate the lidar data such that rigid \ac{ICP} can be applied instead of a motion-compensated implementation. Our method is fast on a single thread, leaving threads available for other important tasks in an autonomy pipeline (e.g., localization, planning, control).


\section{Conclusion} 
  \label{sec:conclusion}
  We presented a continuous-time linear estimator for the 6-\ac{DOF} vehicle velocity using \ac{FMCW} lidar and gyroscope measurements. As our method is linear and does not require data association, it is efficient and capable of operating at an average wall-clock time of 5.64ms for each lidar frame. We demonstrate our method on real-world driving sequences over several kilometers, presenting a compelling trade-off in computation versus accuracy compared to existing state-of-the-art lidar odometry.

As demonstrated in our observability study, and experimentally by Kellner et al. \cite{kellner2014instantaneous} for 3-\ac{DOF} motion, multiple \ac{FMCW} sensors are required for the Doppler measurements to fully constrain the vehicle motion. We plan on using multiple \ac{FMCW} lidars to estimate vehicle motion without the built-in gyroscope. We will further investigate the Doppler measurement bias and work on improving the regression model by introducing more input features (e.g., angle of incidence). We will also incorporate these features into learning a feature-dependant Doppler noise variance.

\appendix
   \subsection{Proof of \autoref{lemma:nullspaceSum}}
    \label{app:proof}
    Assume $\bx\in\nullspace \bA \cap \nullspace \bB$.
    It is straightforward to see that $\bx \in \nullspace{\bA+\bB}$, hence $ \nullspace{\bA+\bB} \supseteq \nullspace \bA \cap \nullspace \bB$.
    Then, assume  $\bx \in \nullspace{\bA+\bB}$, we have
    \vspace{-1em}
    \begin{align*}
    &\bx^T(\bA+\bB)\bx = \underbrace{\bx^T\bA\bx}_{\geq 0} + \underbrace{\bx^T\bB\bx}_{\geq 0} = 0 
    \end{align*}
    Recall that for any symmetric PSD matrix $\bm{M}$, we have $\bm{M}=\bm{S}^T\bm{S}$.
    As such, $\bx^T\bm{M}\bx=0 \Leftrightarrow \left(\bm{S}\bx\right)^T\bm{S}\bx=0 \Rightarrow \bx\in\nullspace{\bm{S}} \Rightarrow \bx\in\nullspace{\bm{M}}$.
    Therefore, $\bx \in \nullspace{\bA} \cap \nullspace{\bB}$ and $ \nullspace{\bA+\bB} \subseteq \nullspace \bA \cap \nullspace \bB$, which concludes the proof.

\section*{Acknowledgments}
We would like to thank the Natural Sciences and Engineering Research Council of Canada (NSERC) and the Ontario Research Fund: Research Excellence (ORF-RE) program for supporting this work.

\bibliographystyle{bib/IEEEtran}
\bibliography{bib/refs, bib/refs_yuchen, bib/IMU_lidar_fusion}

\end{document}